\documentclass[runningheads]{llncs}

\usepackage{eccv}

\usepackage[dvipsnames]{xcolor}

\usepackage{graphicx}
\usepackage{algorithm}
\usepackage{algpseudocode}

\usepackage{amsmath}

\usepackage{amssymb}
\usepackage{booktabs}
\usepackage{bm}
\usepackage{adjustbox}
\usepackage{mathtools}
\usepackage{amsfonts}
\usepackage{array}
\newcolumntype{x}[1]{>{\centering\let\newline\\\arraybackslash\hspace{0pt}}p{#1}}
\usepackage{tikz}
\usepackage{ctable}
\usepackage{float}
\usepackage{enumitem}
\usepackage{dsfont}
\usepackage{xcolor}         
\usepackage{dashrule}       
\usepackage{makecell}       
\usepackage{colortbl}       
\usepackage{multirow}

\usepackage[pagebackref,breaklinks,colorlinks,citecolor=eccvblue]{hyperref}

\definecolor{blueblack}{RGB}{0, 108, 173}
\definecolor{taborange}{RGB}{235, 127, 14}
\definecolor{tabgreen}{RGB}{30, 160, 30}
\definecolor{tabgray}{RGB}{142, 142, 142}
\definecolor{tabpurple}{RGB}{128, 103, 189}
\definecolor{tabred}{RGB}{214, 39, 40}

\DeclareMathOperator*{\argmin}{arg\,min}

\definecolor{sol_light_blue}{RGB}{38, 139, 210}
\definecolor{sol_blue}{RGB}{38, 139, 210}
\definecolor{nord_blue}{RGB}{38, 139, 210}
\definecolor{sol_green}{RGB}{163, 190, 140}
\definecolor{sol_red}{RGB}{220, 50, 47}
\definecolor{nord_red}{RGB}{250, 190, 192}
\definecolor{nord_green}{RGB}{163, 190, 140}

\definecolor{beer_orange}{RGB}{242, 142, 28}

\definecolor{nordblack}{RGB}{46, 52, 64}
\definecolor{nordred}{RGB}{191, 97, 106}
\definecolor{magenta}{RGB}{215, 10, 185}
\definecolor{nordgreen}{RGB}{143, 170, 120}
\definecolor{nordblue}{RGB}{94, 129, 172}
\definecolor{nordpurple}{RGB}{180, 142, 160}

\usepackage[capitalize]{cleveref}
\crefname{section}{Sec.}{Secs.}
\Crefname{section}{Section}{Sections}
\Crefname{table}{Table}{Tables}
\crefname{table}{Tab.}{Tabs.}

\newcommand{\A}{\mathbf{A}}

\newcommand{\h}{\mathbf{h}}
\renewcommand{\H}{\mathbf{H}}

\newcommand{\z}{\mathbf{z}}

\newcommand{\x}{\mathbf{x}}
\newcommand{\y}{\mathbf{y}}

\newcommand{\w}{\mathbf{w}}

\newcommand{\Q}{\mathbf{Q}}
\newcommand{\K}{\mathbf{K}}

\renewcommand{\z}{\mathbf{z}}
\newcommand{\Z}{\mathbf{Z}}

\newcommand{\X}{\mathbf{X}}

\newcommand{\V}{\mathbf{V}}

\renewcommand{\P}{\mathbf{P}}

\renewcommand{\V}{\mathbf{V}}

\newcommand{\M}{\mathbf{M}}

\newtheorem{prop}{Proposition}

\usepackage{eccvabbrv}

\usepackage[accsupp]{axessibility}  

\usepackage{orcidlink}

\begin{document}

\title{Structured Initialization for Attention in\\ Vision Transformers} 

\titlerunning{Structured initialization}

\author{Jianqiao Zheng~\thanks{Corresponding email: jianqiao.zheng@adelaide.edu.au} \and
Xueqian Li \and
Simon Lucey}

\authorrunning{J.~Zheng~\etal}

\institute{AIML, The University of Adelaide \\
\url{https://github.com/osiriszjq/structured_init}}

\maketitle

\begin{abstract}
    The training of vision transformer (ViT) networks on small-scale datasets poses a significant challenge. 
    By contrast, convolutional neural networks (CNNs) have an architectural inductive bias enabling them to perform well on such problems. 
    In this paper, we argue that the architectural bias inherent to CNNs can be reinterpreted as an initialization bias within ViT. 
    This insight is significant as it empowers ViTs to perform equally well on small-scale problems while maintaining their flexibility for large-scale applications. 
    Our inspiration for this ``structured'' initialization stems from our empirical observation that random impulse filters can achieve comparable performance to learned filters within CNNs. 
    Our approach achieves state-of-the-art performance for data-efficient ViT learning across numerous benchmarks including CIFAR-10, CIFAR-100, and SVHN. 
\end{abstract}


\section{Introduction}
\label{sec:intro}

\begin{figure}[t]
    \centering
    {\includegraphics[width=\linewidth]{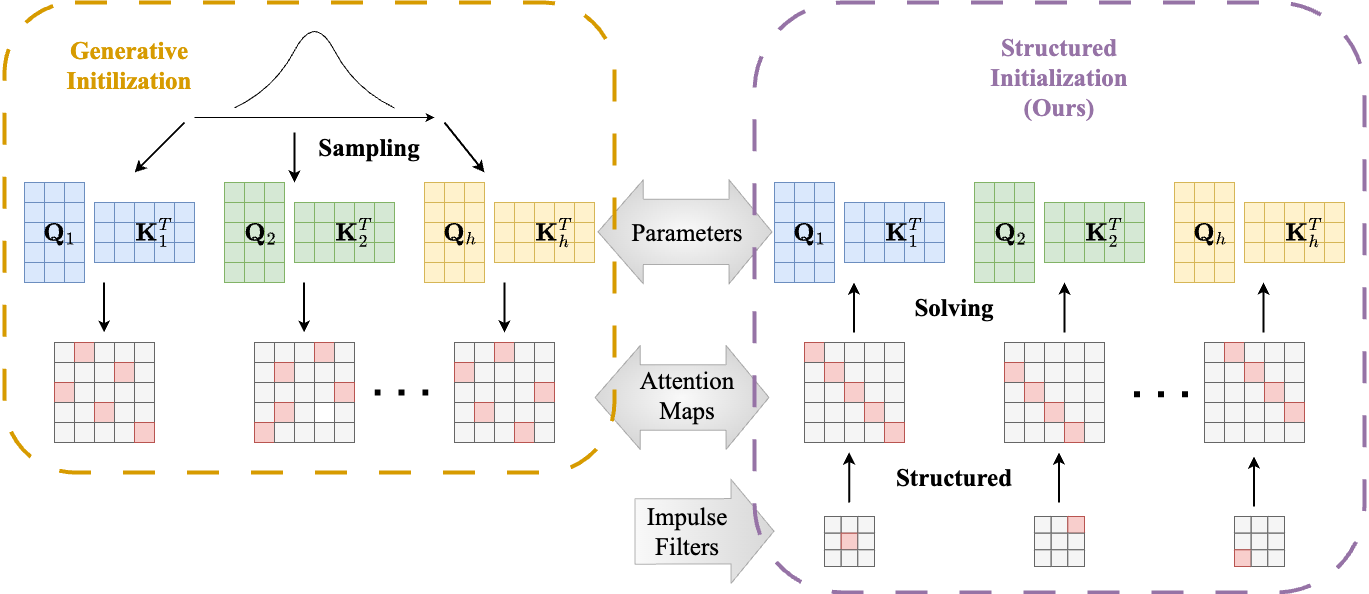}}
    \caption{Illustration comparing conventional generative initialization with structured initialization (ours) strategy for the weights $\mathbf{Q}$ and $\mathbf{K}$ of the self-attention in transformers. 
    Conventional generative initialization involves sampling parameters $\mathbf{Q}$ and $\mathbf{K}$ from certain distributions, such as Gaussian or Uniform, resulting in unstructured initial attention maps.
    In contrast, our structured initialization strategy imposes constraints on the initial structure of the attention maps, specifically requiring them to be random impulse filters.
    The initialization of parameters $\mathbf{Q}$ and $\mathbf{K}$ is computed based on this requirement on attention maps.
    Note that in both attention maps and random impulse filters, the pink cells indicate ones, while the gray cells represent zeros.
    }%
    \vspace{-0.4cm}
    \label{fig:init}%
\end{figure}

Vision Transformer (ViT) networks have shown significant promise when exposed to large amounts of training data.
Their performance, however, is diminished when applied to small-scale datasets where convolutional neural networks (CNNs) still outperform ViTs by a significant margin~\cite{lee2021vision, liu2021efficient}. 
To address this drawback, ViT~\cite{dosovitskiy2020vit} has employed pre-training on larger-scale datasets such as ImageNet~\cite{ILSVRC15}, JFT-300M~\cite{sun2017revisiting},~\etc.
However, this approach has a fundamental limitation. 
One motivation for training visual learning algorithms, such as CNNs, on smaller-scale datasets is the ability to quickly and cost-effectively evaluate their suitability for larger-scale learning tasks. 
Requiring an algorithm to pre-train on a large-scale dataset before training on a smaller-scale dataset essentially undermines this purpose. 
Overcoming this limitation in modern ViTs will make them more accessible and potentially spark new waves of innovation and exploration within the vision community.

Recent advances in CNNs, most notably ConvMixers~\cite{trockman2022patches}, have become increasingly similar in architecture to ViTs. 
Specifically, they split multi-channel convolution into depth-wise and channel-wise convolution. 
Cazenavette~\etal~\cite{cazenavetterethinking} recently demonstrated that random depth-wise (spatial mixing) convolution filter weights can achieve comparable performance to learned weights in ConvMixer and ResNet~\cite{he2016deep} frameworks. 
The only parameters that need to be learned are those associated with channel-wise convolution, commonly referred to as channel mixing in the transformer literature~\cite{tolstikhin2021mlp}. 
This result is particularly intriguing, suggesting that the utility of CNNs lies largely in their convolution structure rather than the exact weights of the filters. 
ViTs still utilize channel-wise convolution for channel mixing but substitute depth-wise convolution with multi-headed attention for spatial mixing.
However, unlike depth-wise convolution, multi-headed attention lacks a predefined structure and instead learns spatial relations from data, which makes it challenging to learn on small-scale datasets.

Expanding upon this insight, we propose to reinterpret the architectural bias found in CNNs into an initialization bias within ViTs. 
Specifically, we provide a theoretical explanation for the effectiveness of random channel-wise convolution filters in ConvMixer.
Furthermore, we demonstrate that utilizing impulse filters yields similarly impressive performance without the need for learning.
While the theoretical equivalence between impulse filters and random filters may appear trivial in ConvMixer, the utilization of impulse filters in ViTs differs substantially from random filters as it inherently builds the softmax structure.
An impulse filter contains only ones and zeros, with each row (matching the softmax dimension) having precisely a single non-zero entry, which can be effectively modeled as a softmax attention matrix.
Based on these observations, we propose that the convolutional inductive bias inherent in CNNs through their architecture can be realized through a structured initialization of the attention maps within ViT, specifically initializing the attention maps within ViT with convolutional matrices of impulse filters.

Traditional initialization strategies typically follow a generative approach, where parameters are sampled from identically independent distributions.
Many works~\cite{he2015delving, liu2022convnet} focus on exploring the suitable distribution for this purpose.
However, the unique architecture of ViT emphasizes the importance not only of the parameters in $\mathbf{Q}$ and $\mathbf{K}$ but also of the attention maps they formalized. 
In contrast to conventional generative approaches, our structured initialization imposes specific structural requirements on the attention maps, controlling $\mathbf{Q}$ and $\mathbf{K}$ without directly constraining trainable parameters (see~\cref{fig:init}).
Additionally, our strategy involves designing different initializations for different attention heads, as the attention map of each head resembles a distinct impulse filter. 
To the best of our knowledge, we are the first to adopt this innovative approach, which embeds some of the architectural principles of CNNs as an initialization bias within ViTs.

\noindent Our paper makes the following contributions:
\begin{itemize}
    \item We provide a theoretical explanation for the effectiveness of random spatial convolution filters, attributing their success to the redundancy in embeddings. 
    Specifically, we argue that the learning of spatial convolution filters can be attributed to channel mixing weights as long as the redundancy condition is satisfied.
    \item We propose to use convolutional structured impulse initialization on the attention maps in ViT.
    Our approach embeds this CNN architectural inductive bias as a structured initialization while preserving the flexibility of transformers in learning attention maps.
    \item We demonstrate state-of-the-art performance for data-efficient ViT learning across various benchmarks including CIFAR-10, CIFAR-100~\cite{krizhevsky2009learning}, and SVHN~\cite{yuval2011reading}. 
    Our approach outperforms the recent mimetic approach~\cite{trockman2023mimetic} and offers a deeper understanding of the ViT initialization.
\end{itemize}

\vspace{-0.1cm}
\section{Related Work}
\label{sec:related_work}
\vspace{-0.1cm}

\noindent \textbf{Convolution as attention.}\,\, 
Since their introduction~\cite{vaswani2017attention,dosovitskiy2020vit}, the relationship between transformers and CNNs has been a topic of immense interest to the vision community. 
Andreoli~\cite{andreoli2019convolution} studied the structural similarities between attention and convolution, bridging them into a unified framework. 
Building on this, Cordonnier~\etal~\cite{Cordonnier2020On} demonstrated that self-attention layers can express any convolutional layers through a careful theoretical construction. 
However, while these studies highlighted the functional equivalence between self-attention in ViTs and convolutional spatial mixing in CNNs, they did not delve into how the inductive bias of ViTs could be adapted or enhanced through this theoretical connection. 
Moreover, the reliance on relative positional encoding, which is not widely used in current mainstream ViT implementations, limited the practical impact of these insights.
In contrast, our work offers a simpler insight: random impulse filter convolution can be effectively approximated by softmax self-attention.

\noindent \textbf{Bias through architecture.}\,\,
Many efforts have aimed to incorporate convolutional inductive bias into ViTs through architectural modifications. 
Dai~\etal~\cite{dai2021coatnet} proposed to combine convolution and self-attention by mixing the convolutional self-attention layers. 
Both Pan~\etal~\cite{pan2021integration} and Li~\etal~\cite{li2023uniformer} introduced hybrid models wherein the output of each layer is a summation of convolution and self-attention. 
Wu~\etal~\cite{wu2021cvt} explored using convolution for token projections within self-attention, while Yuan~\etal~\cite{yuan2021incorporating} demonstrated promising results by inserting a depthwise convolution before the self-attention map as an alternate strategy for injecting inductive bias.
d’Ascoli~\etal~\cite{d2021convit} introduced gated positional self-attention (GPSA) to imply a soft convolution inductive bias.
However, these techniques have a fundamental limitation---they aim to introduce the inductive bias of convolution through architectural choices. 
Our approach, on the other hand, stands out by preserving the architectural flexibility of ViT through a novel CNN-inspired initialization strategy.
Such an approach offers several advantages as it: (i) exhibits data efficiency on small-scale datasets, (ii) retains the architectural freedom to be seamlessly applied to larger-scale datasets, and (iii) gives an alternate theoretical perspective on how the inductive bias of convolution can be applied within transformers.

\noindent \textbf{Bias through initialization.}\,\,
To date, the exploration of applying inductive bias through initialization within a transformer is limited. 
Zhang~\etal~\cite{zhang2022unveiling} posited that the benefit of pre-trained models in ViTs can be interpreted as a more effective strategy for initialization.
Trockman~\etal~\cite{trockman2023mimetic, trockman2022understanding} recently investigated the empirical distributions of self-attention weights, learned from large-scale datasets, and proposed a mimetic initialization strategy. 
While this approach lies between structured and generative initialization, it relies on the simple structure derived from pre-trained large models.
A key difference in our approach is that it does not require offline knowledge of pre-trained networks (mimetic or empirical). 
Instead, our strategy is intrinsically connected to the inductive bias of convolution, without the need to incorporate convolution as an architectural choice.

\begin{figure}[t]
    \centering
    \includegraphics[width=0.7\textwidth]{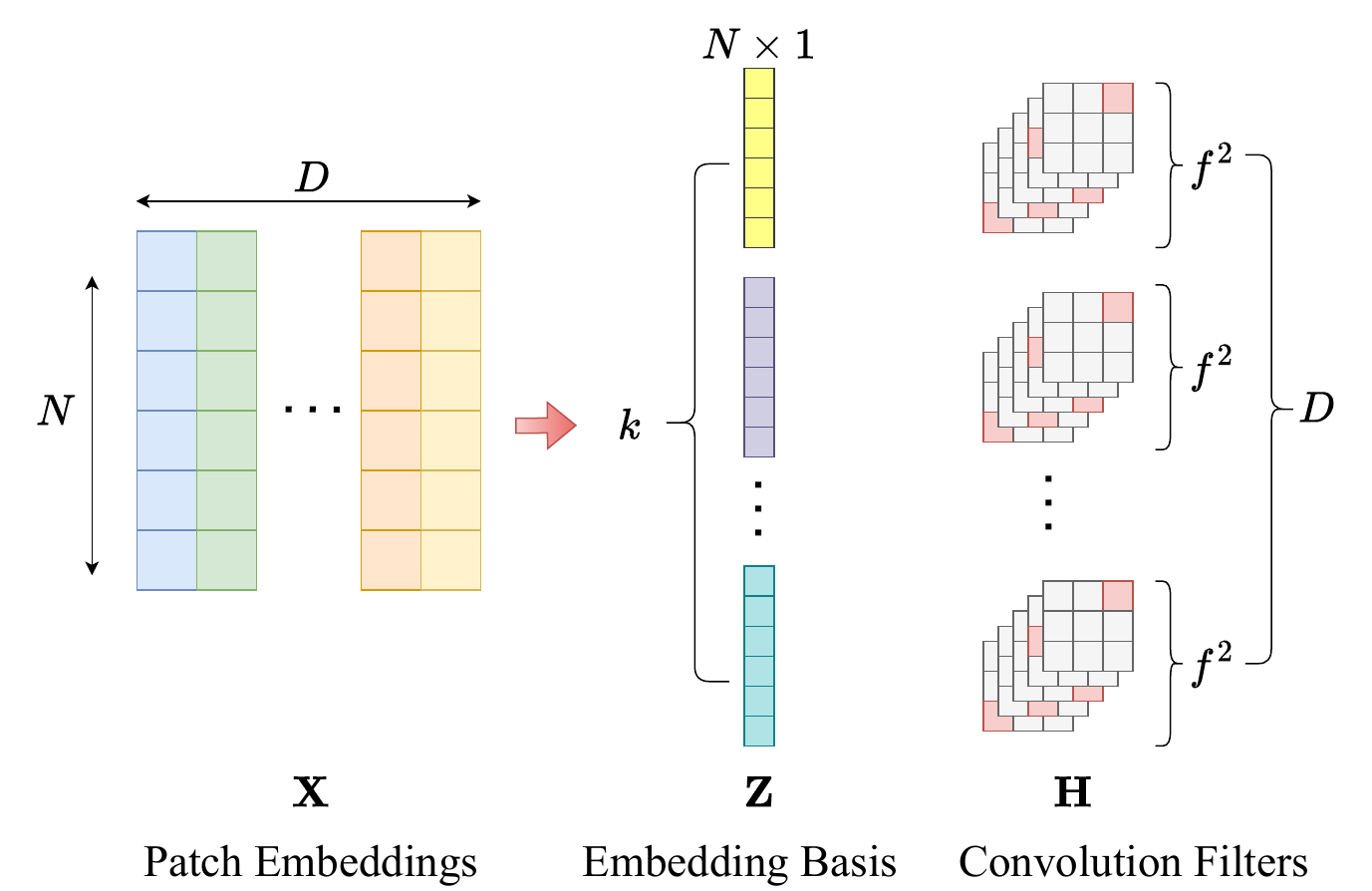}
    \caption{Illustration of why random spatial convolution filters are effective. 
    Patch embeddings $\X\,{\in}\,\mathbb{R}^{N{\times}D}$ are typically rank-deficient and can be approximately decomposed to $k$ basis. 
    Meanwhile, a linear combination of $f^{2}$ linearly independent filters~$\h$ can express any arbitrary filter in the filter space~$\mathbb{R}^{f {\times} f}$. 
    Based on these two observations, we derive the inequality $D\,{\geq}\,k f^2$ in~\cref{prop:convmixer}.}
    \label{fig:sm_conv}
    \vspace{-0.4cm}
\end{figure}

\vspace{-0.1cm}
\section{Preliminaries: Why Random Filters Work?}
\label{sec:sm_conv}
\vspace{-0.1cm}

Cazenavette~\etal~\cite{cazenavetterethinking} recently demonstrated that randomly initialized convolution filters, in networks such as ConvMixer and ResNet, work remarkably well when solely learning the channel mixing parameters. 
However, they failed to offer any insights into the underlying reasons.
In this section, we provide a brief theoretical intuition. 
This finding is significant as it establishes a conceptual link between the architecture of ConvMixer and the initialization of ViT, offering a deeper understanding of desired properties for spatial mixing matrices.
For clarity and simplicity, we have omitted activations (\eg, GeLU, ReLU,~\etc), bias, batch normalization, and skip connections in our formulation.

\begin{remark}
\label{remark:redundency}
Let us define the patch embeddings or intermediate layer outputs in ConvMixer as $\X\,{=}\,[\x_{1},\x_{2},\dots,\x_{D}]$, where~$D$ is the number of channels and~$N$ is the number of pixels in the vectorized patch~$\x \,{\in}\, \mathbb{R}^{N}$. 
An interesting observation is the rank (stable rank, defined as $\sum \sigma^{2} / \sigma_{max}^{2}$) of $\X$ is consistently much smaller than the minimum dimension $\mbox{min}(N,D)$ of $\X$, which indicates a significant amount of redundancy in $\X$.
\end{remark}

Let us define a 2D convolution filter as~$\h \,{\in}\, \mathbb{R}^{f {\times} f}$. 
In general, this kernel can be represented as a convolutional matrix~$\H \,{\in}\, \mathbb{R}^{N {\times} N}$, such that~$\h \,{*}\, \x \,{=}\, \H \x$, where~$*$ is the convolutional operator. 
Let $\w\,{=}\, [w_{1},w_{2},\dots,w_{D}]^{T}\,{\in}\, \mathbb{R}^{D {\times} 1}$ be the channel mixing weights for one output channel and $\H_{1}, \H_{2},\dots,\H_{D}$ are the corresponding spatial convolution filters for each channel. 
Therefore, the result~$\y \,{\in}\, \mathbb{R}^{N}$ after spatial and channel mixing can be represented as,
\begin{equation}
    \y = \sum_{i=1}^D w_{i}\H_{i}\x_{i},
    \label{equ:spm_chm_02}
\end{equation}
With~\cref{remark:redundency}, we can suppose the rank of $\X \,{\approx}\, \Z \A$ is $k$, where~$\Z \,{=}\, [\z_{1},\ldots,z_{k}]$ and $k \,{\ll}\,D$, as illustrated in~\cref{fig:sm_conv}.
We then obtain 
\begin{equation}
\begin{aligned}
    \y \approx\sum_{i=1}^D\sum_{j=1}^{k}  w_{i}a_{ji}\H_{i}\z_{j}=\sum_{j=1}^{k}\Tilde{\H}_{j}\z_{j},
\end{aligned}
    \label{equ:spm_chm_z}
\end{equation}
where~$a_{ji}$ refers to the row~$j$, column~$i$ element of~$\A$, and $\Tilde{\H}_{j}\,{=}\,\sum_{i=1}^D  w_{i}\,a_{ji}\,\H_{i}$.

Remember that a linear combination of $f^{2}$ linearly independent filters~$\h$ can express any arbitrary filter in filter space~$\mathbb{R}^{f {\times} f}$, where $\h$ serves as the basis.
Consequently, any desired $\Tilde{\H}_{1}, \Tilde{\H}_{2},\dots,\Tilde{\H}_{D}$ can be achieved by only learning the channel mixing weights $\w$.
Therefore, we obtain the following proposition.
\begin{prop}
\label{prop:convmixer}
    In a ConvMixer block composed of a spatial mixing layer and a channel mixing layer, suppose $D$ is the number of channels, $k$ is the rank of input $\X$, and $f{\times}f$ is the number of convolution filters basis, then any possible output for $f {\times} f$ filters can be achieved by \textbf{only} learning the channel mixing weights, as long as
    \begin{equation}
        D \geq k f^2.
        \label{equ:conv_condition}
    \end{equation}
\end{prop}
Although this derivation simplifies by omitting activations, batch normalizations, and skip connections, it offers an intuition on how solely learning channel mixing can be sufficient for achieving reasonably good performance. 
Furthermore, it reveals that the convolution filters do not necessarily need to be strictly unique from each other. 
As long as we have $f^{2}$ linearly independent convolution filters, each with at least $k$ copies, such as a random impulse filter, the model should still perform well. 
Additionally, this configuration can be viewed as having $f^{2}$ heads, with each head having an embedding dimension of $k$ in multi-head self-attention.

\vspace{-0.1cm}
\section{Method}
\label{sec:method}
\vspace{-0.1cm}

ConvMixer and ViT share most components in their architectures. 
The gap in their performance on small-scale datasets stems from their architectural choices regarding spatial mixing matrix.
Although depthwise convolution (ConvMixer) and multi-head self-attention (ViT) may appear distinct at first glance, their underlying goal remains the same: to identify spatial patterns indicated by the spatial mixing matrix.
The spatial mixing step can be expressed as 
\begin{equation}
\label{equ:spatial_mixing}
    \mathbf{x} \leftarrow \M \mathbf{x},
\end{equation}
where~$\X \,{=}\, [\x_{1},\ldots,\x_{D}]$,~$\x \,{\in}\, \mathbb{R}^{N}$ are the patch embeddings or intermediate outputs, and~$\M$ is the~$N {\times} N$ spatial mixing matrix.

In ConvMixer, there exists a strong assumption that the pattern only appears in the local area defined by the convolutional kernel size. 
Therefore, the spatial mixing matrix~$\M$ can be viewed as the convolutional matrix~$\H$ in~\cref{equ:spm_chm_02}. 
In particular, for each channel $\x_{1},\ldots,\x_{D}$, the corresponding spatial mixing matrix $\M_{1}, \M_{2},\dots,\M_{D}$ is $\H_{1}, \H_{2},\dots,\H_{D}$.

In ViT, this assumption is noticeably weaker. 
The pattern is learned from data, based on the similarity of projected input. 
Therefore, the matrix~$\M$ for multi-head self-attention can be expressed as follows:
\begin{equation}
\label{equ:attention}
\M_{i}^{\hphantom{T}} = \mbox{softmax}(\X \mathbf{Q}_{i}^{\hphantom{T}} \mathbf{K}_{i}^{T} \mathbf{X}^{T}),
\end{equation}
where $\mathbf{Q}_{i}^{\hphantom{T}},\mathbf{K}_{i}^{\hphantom{T}} \,{\in}\, \mathbb{R}^{D {\times} K}$ denote the attention weight matrices, $i\,{=}\,1,\ldots,h$, and~$K$ represents the feature dimension in each head, typically set to $D/h$, with $h$ being the number of heads. 
In summary, for channels $\x_{1},\ldots,\x_{D}$, the spatial mixing matrices are $\M_{1}, \M_{2},\dots,\M_{h}$ where each $\M_{i}$ is shared by $D/h$ channels. 
It is similar to our findings in ConvMixer, except that there are in total $h$ unique spatial mixing matrices and each has $D/h$ copies. 
Recall our discussion in~\cref{prop:convmixer}, where we noted that the filter matrices $\H_{1}, \H_{2},\dots,\H_{D}$ need not to be unique from each other. 
Leveraging this insight, we propose to initialize the attention map for each head in ViT with a convolutional filter matrix $\H$.
Our approach integrates the architectural bias inherent in CNNs into the initialization of attention maps within ViT. 
For clarity and brevity, the following discussions will focus only on one head of multi-head self-attention.

\subsection{Structured Convolutional Initialization}
\label{sec:sci}
Based on our observation in ConvMixer and ViT, we propose to use a structured convolutional initialization strategy to
incorporate prior knowledge into the initialization of the attention map, imposing a convolutional structure on it:
\begin{equation}
\label{equ:init}
\M_{\text{init}}^{\hphantom{T}} = \mbox{softmax}(\X \mathbf{Q}_{\text{init}}^{\hphantom{T}} \mathbf{K}_{\text{init}}^{T} \mathbf{X}^{T}) \approx \H.
\end{equation}

\noindent\textbf{Why using impulse filters?}\,\,
Usually, random convolution filters contain both positive and negative values, while the output of the softmax function is always positive. 
One straightforward option is to use random positive convolution filters with a normalized sum of one, following the property of softmax.
However, this approach often proves inefficient as the patterns may be too complicated for a softmax function to handle. 
Tarzanagh~\etal~\cite{tarzanagh2023transformers} found that the softmax attention map functions as a feature selection mechanism, and typically tends to select a single related feature.
In convolution filters, this softmax attention map can be viewed as an impulse filter.
According to~\cref{prop:convmixer}, there exists no distinction in the choices of different filters, as long as these filters form the basis for the $f^{2}$ space. 
In conclusion, when initializing a softmax attention map, the most straightforward and suitable choice is random impulse convolution filters.

\noindent\textbf{Pseudo input.}\,\,
The advantage of self-attention is that its spatial mixing map is learned from data. 
However, during the initialization phase, there is no prior information about the input.
To address this problem, we use absolute sinusoidal positional encoding $\P$ as pseudo input, replacing the original $\X$.
In addition, we explore alternative pseudo inputs such as random inputs sampled from Gaussian or Uniform distribution, a combination of random inputs and positional encoding,~\etc. 
The ablation study of different pseudo inputs is presented in~\cref{sec:exp_input}.
While some of these choices may yield promising results, the simplest and most reasonable approach remains the utilization of positional encoding.

With these two adaptions,~\cref{equ:init} becomes
\begin{equation}
\label{equ:init_p}
\M_{\text{init}}^{\hphantom{T}} = \mbox{softmax}(\P \mathbf{Q}_{\text{init}}^{\hphantom{T}} \mathbf{K}_{\text{init}}^{T} \mathbf{P}^{T}) \approx \H_{\text{impulse}}^{\hphantom{T}}.
\end{equation}

\subsection{Solving $\Q_{\text{init}}^{\hphantom{T}} $ and $\K_{\text{init}}$}\label{sec:solving_qk}

There exist numerous approaches to solve~\cref{equ:init_p} for $\Q_{\text{init}}^{\hphantom{T}}$ and $\K_{\text{init}}^{\hphantom{T}}$ with known $\H_{\text{impulse}}^{\hphantom{T}}$ and $\P$. 
In mimetic initialization~\cite{trockman2023mimetic}, $\mathbf{Q}_{\text{init}}^{\hphantom{T}} \mathbf{K}_{\text{init}}^{T}$ is initialized, and SVD is utilized to solve $\mathbf{Q}_{\text{init}}^{\hphantom{T}}$ and $\mathbf{K}_{\text{init}}^{\hphantom{T}}$. 
While a similar SVD-based approach could be employed in our scenario---despite we intend to initialize the softmax attention map, it is found to be ineffective due to the large error resulting from the pseudo-inverse of $\mathbf{P}$ and low-rank approximation.
Consequently, we opt not to pursue an analytical solution but rather employ a simple optimization to obtain $\mathbf{Q}_{\text{init}}$ and $\mathbf{K}_{\text{init}}$. 
This approach also addresses concerns regarding scale and layer normalization in the attention mechanism.

\begin{algorithm}
\caption{Convolutional structured impulse initialization for ViT}\label{alg:cap}
\begin{algorithmic}
\Require $\P,\,f$ \Comment Positional encoding, convolution filter size
\Ensure $\Q_{\text{init}}^{\hphantom{T}},\,\K_{\text{init}}^{\hphantom{T}}$ \Comment Initialization of attention parameters
\State $N,D \gets \text{shape of }\P$
\State $\H_{\text{impulse}}^{\hphantom{T}} \gets ImpulseConvMatrix(N,f)$ \Comment Build 2D impulse convolution matrix
\State $\Tilde{\X} \gets LayerNorm(\P)$ \Comment Get pseudo input
\State $\sigma \gets \frac{1}{\sqrt{D/h}}$ \Comment Scale in attention
\State $\Q_{\text{init}}^{\hphantom{T}},\,\K_{\text{init}}^{\hphantom{T}} \gets Parameters(\cdot)$ \Comment Random initialized before optimization
\For{$i \gets 1, max\_iter$}
\State $\Hat{\H}_{\text{impulse}}^{\hphantom{T}} \gets \mbox{softmax}(\sigma\Tilde{\X} \Q_{\text{init}}^{\hphantom{T}} \K_{\text{init}}^{T} \Tilde{\X}^{T})$
\State $Loss \gets \mbox{MSE}(\Hat{\H}_{\text{impulse}}^{\hphantom{T}},\H_{\text{impulse}}^{\hphantom{T}})$
\State Compute gradients and update $\Q_{\text{init}}^{\hphantom{T}}$ and $\K_{\text{init}}^{\hphantom{T}}$
\EndFor \\
\Return $\Q_{\text{init}}^{\hphantom{T}},\,\K_{\text{init}}^{\hphantom{T}}$
\end{algorithmic}
\end{algorithm}

The pseudo code for our initialization strategy is shown in~\cref{alg:cap}. 
In the first step, we compute the attention map $\H_{\text{impulse}}^{\hphantom{T}}$ based on the 2D impulse convolution matrix.
The pseudo input $\Tilde{\X}$ is then computed through the absolute positional encoding $\P$.
Note that the pseudo input $\Tilde{\X}$ remains constant throughout the entire optimization process without requiring re-sampling.
Additionally, the constant scale $\sigma$, and any normalization techniques such as layer normalization or batch normalization remain consistent with those utilized in ViT.

To optimize $\Q_{\text{init}}^{\hphantom{T}}$ and $\K_{\text{init}}^{\hphantom{T}}$, our objective function is defined as
\begin{align}
    \argmin_{\Q_{\text{init}}^{\hphantom{T}}, \K_{\text{init}}^{\hphantom{T}}} \frac{1}{N^2} \left\Vert \H_{\text{impulse}}^{\hphantom{T}} - \mbox{softmax} \left(\sigma\Tilde{\X} \Q_{\text{init}}^{\hphantom{T}} \K_{\text{init}}^{T} \Tilde{\X}^{T} \right) \right\Vert^2_F,
    \label{eq:initial_network}
\end{align}
where $\Tilde{\X}$ is the normalized pseudo input, and $\Q_{\text{init}}^{\hphantom{T}}$ and $\K_{\text{init}}^{\hphantom{T}}$ can be random initialized before optimization.
We then optimize for $max\_iter \,{=}\,10,000$ epochs using Adam optimizer~\cite{kingma2014adam} with a learning rate of $1e^{-4}$ using mean squared error (MSE) loss. 
It is worth noting that this optimization does not count as a pre-training step since no real data is involved.
Rather, our optimization algorithm serves as a surrogate for the SVD solver, converging in just a few seconds ($\sim$5s).

\vspace{-0.1cm}
\section{Experiments and Analysis}
\label{sec:exp}
\vspace{-0.1cm}
\vspace{-0.1cm}
\subsection{Settings}
\label{sec:exp_settings}
\vspace{-0.1cm}
\noindent\textbf{Dataset.}\,\,
We evaluate our structured initialization strategy on the small-scale datasets CIFAR-10, CIFAR-100~\cite{krizhevsky2009learning}, SVHN~\cite{yuval2011reading}.
Additionally, we test our model on a large-scale ImageNet-1K~\cite{imagenet_cvpr09} dataset.
Furthermore, in validating our theory on ConvMixer, we conduct all related experiments on CIFAR-10.

\noindent\textbf{Models.}\,\,
Our experiments primarily focus on the tiny ViT model, namely ViT-T~\cite{dosovitskiy2020vit}. 
Instead of using the classification token and a learnable positional encoding as defined in ViT, we use the average global pooling and a sinusoidal absolute positional encoding.
In general, these small tweaks will not compromise the performance of ViTs.
On the contrary, as shown in~\cref{tab:modified_vit}, these two modifications lead to improved performance on the CIFAR-10 dataset. 
Henceforth, all the following experiments use this configuration. 
The default architecture of ViT-T includes a depth of 12, an embedding dimension of 192, and 3 heads.

\begin{table}[t]
\caption[]{Classification accuracy(\%) of ViT-T with various basic settings on CIFAR-10. }
\centering
\begin{adjustbox}{width=0.6\linewidth}
\begin{tabular}
{@{}clx{0.3\linewidth}x{0.25\linewidth}@{}}
\toprule
& \thead{\normalsize Model} & \thead{\normalsize Classification Token} & \thead{\normalsize Average Pooling }  \\
\midrule
& Learnable PE & 81.23 & 82.23\\
& Sinusoidal PE & 83.17 & \textbf{85.30}  \\
\bottomrule
\end{tabular}
\label{tab:modified_vit}
\end{adjustbox}
\vspace{-0.4cm}
\end{table}

\noindent\textbf{Training.}\,\,
We utilize the PyTorch Image Models (timm)~\cite{rw2019timm} to train all ViT models. 
We employ a simple random augmentation strategy from~\cite{cubuk2020randaugment} for data augmentation. 
Our models were trained with a batch size of 512 using the AdamW \cite{loshchilov2018decoupled} optimizer, with a learning rate of $10^{-3}$ and weight decay set to 0.01, for 200 epochs. 
Note that all experiments were conducted on the Tesla V100 SXM3 with 32GB memory.

\noindent\textbf{Initialization.}\,\,
Considering that the number of heads in ViTs is typically small, we utilized both $3{\times}3$ (Imp.-3) and $5{\times}5$ (Imp.-5) filters for our convolutional structured impulse initialization method.
We compare our method with Pytorch default initialization (Kaiming Uniform~\cite{he2015delving}), timm default initialization (Trunc Normal), and mimetic initialization (Mimetic~\cite{trockman2023mimetic}).

\vspace{-0.2cm}
\subsection{Results Across Datasets}
\label{sec:exp_dataset}
\vspace{-0.1cm}
In~\cref{tab:dataset}, we present the results of five different methods across four datasets. 
For ImageNet-1K, we follow the training settings defined in ConvMixer~\cite{trockman2022patches}, training all models for 300 epochs.
Our proposed methods, both Imp.-3 and Imp.-5, demonstrate comparable---if not superior---performance compared to mimetic initialization. 
Particularly on smaller-scale datasets like CIFAR-10, CIFAR-100, and SVHN, known to pose challenges for ViT models, our method consistently exhibits $2\%$ to $4\%$ improvement compared to Trunc Normal. 
Notably, our method maintains to perform well on large-scale datasets like ImageNet-1K, which shows that our structured initialization keeps the flexibility of the attention map even when learning from large-scale data.

\vspace{-0.2cm}
\subsection{Larger Models}
\label{sec:exp_larger}
\vspace{-0.1cm}
Although our method demonstrates impressive performance when training ViT-T on small-scale datasets, the model ViT-T only has 3 heads, which falls short of the requirements defined in~\cref{prop:convmixer}. 
To better showcase the advantage of using our initialization strategy, we increase the number of heads to 8 in ViT-T, denoted as ViT-T/h8.
In addition to the experiments with ViT-T, we also tested our method on the small ViT model (ViT-S).
The configuration of ViT-S includes an embedding dimension of 384, a depth of 12, and 6 heads, denoted as ViT-S/h6.
Furthermore, we increase the number of heads to 16 and denote this model as ViT-S/h16. 
The results on CIFAR-100 are shown in~\cref{tab:larger_cifar100}.

\begin{table}[t]
\caption[]{Classification accuracy(\%) of ViT-T using different initialization methods on CIFAR-10, CIFAR-100, SVHN and ImageNet-1K.
{\color{tabred}Red} number indicates accuracy decrease, and {\color{tabgreen}green} number indicates an increase in accuracy.
Note that we compare the performance to the Trunc Normal initialization method (shaded in gray).
}
\centering
\begin{adjustbox}{width=0.95\linewidth}
\begin{tabular}
{@{}llx{0.2\linewidth}x{0.2\linewidth}x{0.2\linewidth}x{0.2\linewidth}}
\toprule
& \thead{\normalsize Method} & \thead{\normalsize CIFAR-10} & \thead{\normalsize CIFAR-100} & \thead{\normalsize SVHN}  & \thead{\normalsize ImageNet-1K} \\
\midrule
& Kaiming Uniform~\cite{he2015delving} 
& 86.36{\color{tabred}~~2.27$\downarrow$} 
& 63.50{\color{tabred}~~3.00$\downarrow$} 
& 94.51{\color{tabgreen}~~1.31$\uparrow$} 
& 74.11{\color{tabgreen}~~0.69$\uparrow$} \\
\rowcolor{tabgray!30!} & Trunc Normal 
& 88.63\hphantom{{\color{tabred}~~2.27$\downarrow$}}
& 66.50\hphantom{{\color{tabred}~~2.27$\downarrow$}}
& 93.20\hphantom{{\color{tabred}~~2.27$\downarrow$}}
& 73.42\hphantom{{\color{tabred}~~2.27$\downarrow$}} \\
& Mimetic~\cite{trockman2023mimetic} 
& \underline{91.16}{\color{tabgreen}~~2.53$\uparrow$}
& \underline{70.40}{\color{tabgreen}~~3.90$\uparrow$}
& \textbf{97.53}{\color{tabgreen}~~4.33$\uparrow$}~~
& \underline{74.34}{\color{tabgreen}~~0.92$\uparrow$} \\
& Ours (Imp.-3) 
& \textbf{91.62}{\color{tabgreen}~~2.99$\uparrow$}~~
& 68.81{\color{tabgreen}~~2.31$\uparrow$}
& 97.21{\color{tabgreen}~~4.01$\uparrow$}
& 74.24{\color{tabgreen}~~0.82$\uparrow$} \\
& Ours (Imp.-5) 
& 90.67{\color{tabgreen}~~2.04$\uparrow$}
& \textbf{70.46}{\color{tabgreen}~~3.96$\uparrow$}~~
& \underline{97.23}{\color{tabgreen}~~4.03$\uparrow$} 
& \textbf{74.40}{\color{tabgreen}~~0.98$\uparrow$}~~ \\
\bottomrule
\end{tabular}
\label{tab:dataset}
\end{adjustbox}
\end{table}

\begin{table}[t]
\caption[]{Classification accuracy(\%) of ViT-T/h3, ViT-T/h8, ViT-S/h6 and ViT-S/h16 using different initialization methods on CIFAR-100.
{\color{tabred}Red} number indicates accuracy decrease, and {\color{tabgreen}green} number indicates an increase in accuracy.
Note that we compare the performance to the Trunc Normal initialization method (shaded in gray).
}
\centering
\begin{adjustbox}{width=0.95\linewidth}
\begin{tabular}
{@{}llx{0.2\linewidth}x{0.2\linewidth}x{0.2\linewidth}x{0.2\linewidth}}
\toprule
& \thead{\normalsize Method} & \thead{\normalsize ViT-T/h3} & \thead{\normalsize ViT-T/h8} & \thead{\normalsize ViT-S/h6}  & \thead{\normalsize ViT-S/h16} \\
\midrule
& Kaiming Uniform~\cite{he2015delving} 
& 63.50{\color{tabred}~~3.00$\downarrow$}
& 63.09{\color{tabred}~~3.39$\downarrow$}
& 66.06{\color{tabred}~~0.75$\downarrow$} 
& 64.61{\color{tabred}~~2.64$\downarrow$} \\
\rowcolor{tabgray!30!} & Trunc Normal 
& 66.50\hphantom{{\color{tabred}~~3.00$\downarrow$}}
& 66.48\hphantom{{\color{tabred}~~3.00$\downarrow$}} 
& 66.81\hphantom{{\color{tabred}~~3.00$\downarrow$}} 
& 67.25\hphantom{{\color{tabred}~~3.00$\downarrow$}} \\
& Mimetic~\cite{trockman2023mimetic} 
& \underline{70.40}{\color{tabgreen}~~3.90$\uparrow$}
& 69.93{\color{tabgreen}~~3.45$\uparrow$} 
& \underline{73.86}{\color{tabgreen}~~7.05$\uparrow$}
& 72.72{\color{tabgreen}~~5.47$\uparrow$} \\
& Ours (Imp.-3) 
& 68.81{\color{tabgreen}~~2.31$\uparrow$} 
& \underline{70.79}{\color{tabgreen}~~4.31$\uparrow$} 
& \textbf{75.97}{\color{tabgreen}~~9.16$\uparrow$}~~ 
& \textbf{75.40}{\color{tabgreen}~~8.15$\uparrow$}~~ \\
& Ours (Imp.-5) 
& \textbf{70.46}{\color{tabgreen}~~3.96$\uparrow$}~~ 
& \textbf{70.86}{\color{tabgreen}~~4.38$\uparrow$}~~
& 73.49{\color{tabgreen}~~6.68$\uparrow$} 
& \underline{74.27}{\color{tabgreen}~~7.02$\uparrow$} \\
\bottomrule
\end{tabular}
\label{tab:larger_cifar100}
\end{adjustbox}
\vspace{-0.4cm}
\end{table}

\begin{table}[t]
\caption[]{Classification accuracy(\%) of different pseudo input on CIFAR-10.
{\color{tabgreen}Green} shaded row indicates the best choice of pseudo inputs when achieving the best average accuracy.
``PE'' denotes positional encoding, and ``G'' represents random sampling from the Gaussian distribution.

}
\centering
\begin{adjustbox}{width=\linewidth}
\begin{tabular}
{@{}x{0.08\linewidth}x{0.12\linewidth}cx{0.09\linewidth}x{0.09\linewidth}cx{0.09\linewidth}x{0.09\linewidth}cx{0.09\linewidth}x{0.09\linewidth}cx{0.09\linewidth}x{0.09\linewidth}cx{0.09\linewidth}@{}}
\toprule
\multicolumn{2}{c}{\thead{\normalsize Pseudo Input}} 
& \multicolumn{6}{c}{\thead{\normalsize Same $\Q_{\text{init}}^{\hphantom{T}}$ and $\K_{\text{init}}^{\hphantom{T}}$}} 
& \multicolumn{6}{c}{\thead{\normalsize Different $\Q_{\text{init}}^{\hphantom{T}}$ and $\K_{\text{init}}^{\hphantom{T}}$}} 
&& \multirow{3}{*}{\thead{\normalsize Avg.}} \\ 
\cmidrule{1-2}\cmidrule{4-8}\cmidrule{10-14}
\multirow{2}{*}{\begin{tabular}[c]{@{}c@{}}First\\ Layer\end{tabular}} 
& \multirow{2}{*}{\begin{tabular}[c]{@{}c@{}}Following\\ Layers\end{tabular}} 
&& \multicolumn{2}{c}{ViT-T/h3} 
&& \multicolumn{2}{c}{ViT-T/h8} 
&& \multicolumn{2}{c}{ViT-T/h3} 
&& \multicolumn{2}{c}{ViT-T/h8} 
& \\
\cmidrule{4-5}\cmidrule{7-8}\cmidrule{10-11}\cmidrule{13-14}
& & &Imp.-3 & Imp.-5 && Imp.-3 & Imp.-5 && Imp.-3 & Imp.-5 && Imp.-3 & Imp.-5 & \\ 
\hline
\rowcolor{tabgreen!30!}
PE   & PE   && 90.75 & 90.22 && 90.39 & 90.24 && 89.90 & 90.18 && 90.19 & 91.24 && \textbf{90.39} \\
\hline
PE   & G   && 87.38 & 87.95 && 87.70 & 87.93 && 87.18 & 87.49 && 87.11 & 86.66 && 87.43 \\
PE   & PE+G && 89.69 & 89.27 && 90.15 & 89.38 && 89.73 & 88.90 && 89.62 & 89.09 && 89.48 \\
G   & PE   && 90.37 & 89.54 && 90.44 & 90.11 && 89.81 & 90.57 && 90.49 & 90.84 && 90.27 \\
G   & G   && 86.15 & 86.14 && 86.59 & 85.97 && 86.76 & 86.71 && 86.86 & 85.99 && 86.40 \\
G   & PE+G && 90.20 & 89.85 && 90.21 & 89.60 && 90.00 & 89.71 && 89.61 & 89.46 && 89.83 \\
PE+G & PE   && 90.23 & 89.81 && 90.10 & 89.98 && 89.80 & 90.28 && 90.59 & 90.51 && 90.16 \\
PE+G & G   && 88.02 & 88.31 && 87.59 & 87.94 && 87.57 & 87.25 && 86.73 & 86.38 && 87.47 \\
PE+G & PE+G && 90.01 & 89.69 && 90.01 & 89.66 && 88.79 & 89.36 && 89.56 & 88.98 && 89.51 \\ 
\bottomrule
\end{tabular}
\label{tab:pseudo_input}
\end{adjustbox}
\vspace{-0.6cm}
\end{table}

As the model size increases, particularly with a higher number of heads, our initialization method demonstrates improved and more stable performance, bringing a larger gap between other initialization methods. 
This performance increase proves our theory (see~\cref{prop:convmixer}) regarding the expressibility of spatial mixing matrix: more heads provide more linearly independent filters.
For instance, when the number of heads is 3, as in ViT-T/h3, each layer contains only 3 unique ``filters'' with each filter having $192\,{/}\,3\,{=}\,64$ copies. 
While the number of copies is sufficient, having only 3 unique ``filters'' is inadequate for forming the filter basis, even for a $3\,{\times}\,3$ random impulse filter.

As we increase the number of heads, we observe an adequate improvement in the performance of our method.
However, maintaining a constant embedding dimension while increasing the number of heads leads to fewer copies per head.
While this may not present a significant issue in ConvMxier as long as the number of copies exceeds the rank of the inputs, a notable challenge arises with multi-head attention: the dimensionality of $\mathbf{Q}$ and $\mathbf{K}$ will decrease to $D\,{/}\,h$ as the number of heads $h$ increases. 
Consequently, the rank of $\mathbf{Q}$ and $\mathbf{K}$ diminishes considerably, making it more challenging for the low-rank approximation $\mathbf{Q}\mathbf{K}^{T}$ to learn an effective attention map.

This phenomenon may explain why the Kaiming Uniform and Trunc Normal methods occasionally exhibit inferior performance as the number of heads increases.
For the mimetic initialization, the situation is potentially more problematic, as it utilizes SVD to solve for a low-rank $\mathbf{Q}$ and $\mathbf{K}$. 
As the number of heads increases, resulting in a lower rank, the approximation error grows, further deviating the actual $\mathbf{Q}\mathbf{K}^{T}$ from the anticipated value.
In contrast, our initialization strategy employs an iterative optimization method, which helps mitigate errors arising from low-rank approximations. 
Consequently, our method benefits more when applied with a larger number of heads.

\vspace{-0.2cm}
\subsection{Pseudo Input}
\label{sec:exp_input}
\vspace{-0.2cm}
It is important to note that no real data is involved in our initialization strategy, distinguishing it from pre-training methods.
Incorporating real data into the initialization will unnecessarily complicate the optimization of attention parameters, thereby increasing computational complexity.
Therefore, we use pseudo input to optimize attention parameters $\Q_{\text{init}}^{\hphantom{T}}$ and $\K_{\text{init}}^{\hphantom{T}}$ using gradient descent.
Among various options for pseudo input, sinusoidal absolute positional encoding (PE) proves to be a good choice considering that (1) it remains independent of data, and (2) it inherently embeds the spatial information. 
Nonetheless, alternative choices for pseudo inputs are worth exploring.

Here, we test different pseudo inputs, including random inputs sampled from Gaussian or Uniform distributions.  
As discussed in \cref{sec:solving_qk}, these pseudo inputs are sampled once and kept fixed during the optimization process.
We show this ablation results in \cref{tab:pseudo_input}. 
Note that we only include results for Gaussian distributions (G) that sampled from zero mean and a standard deviation of 0.5, truncated at [-2, 2].
Results for Uniform distributions are included in the supplementary material.
Apart from using the same pseudo input for all network layers, we also explored using different pseudo inputs for individual layers.

During the training of the ViT model, the inputs to different network layers differ in that the input to the first layer is a patch embedding of the image along with a positional encoding, whereas the input to the following layers is the output feature embedding obtained from the previous layer.
To accommodate these differences in network layers, we explore using different pseudo inputs for different layers.
Notably, we treat the first layer separately from the following layers.
Additionally, we explore two configurations regarding the attention parameters: one where the same optimized parameters are utilized across the entire network, and another where these parameters are optimized separately for each layer. These configurations are labeled as ``same'' and ``different'' for $\Q_{\text{init}}^{\hphantom{T}}$ and $\K_{\text{init}}^{\hphantom{T}}$. 
Consequently, we test different pseudo inputs in eight different configurations with ViT-T/h3 and ViT-T/h8.

From~\cref{tab:pseudo_input}, it is evident that using PE as pseudo input yields the best performance. 
Switching from all PEs to PE+G results in a 1\%-3\% performance drop, while using random inputs alone leads to the poorest performance, with accuracy dropping by 4\%.
Additionally, substituting PE with alternative pseudo inputs results in worse performance compared to modifications in the first layer, since there are 11 ``following layers'' in ViT-T.

\begin{figure}[t]
    \centering
    \includegraphics[width=\linewidth]{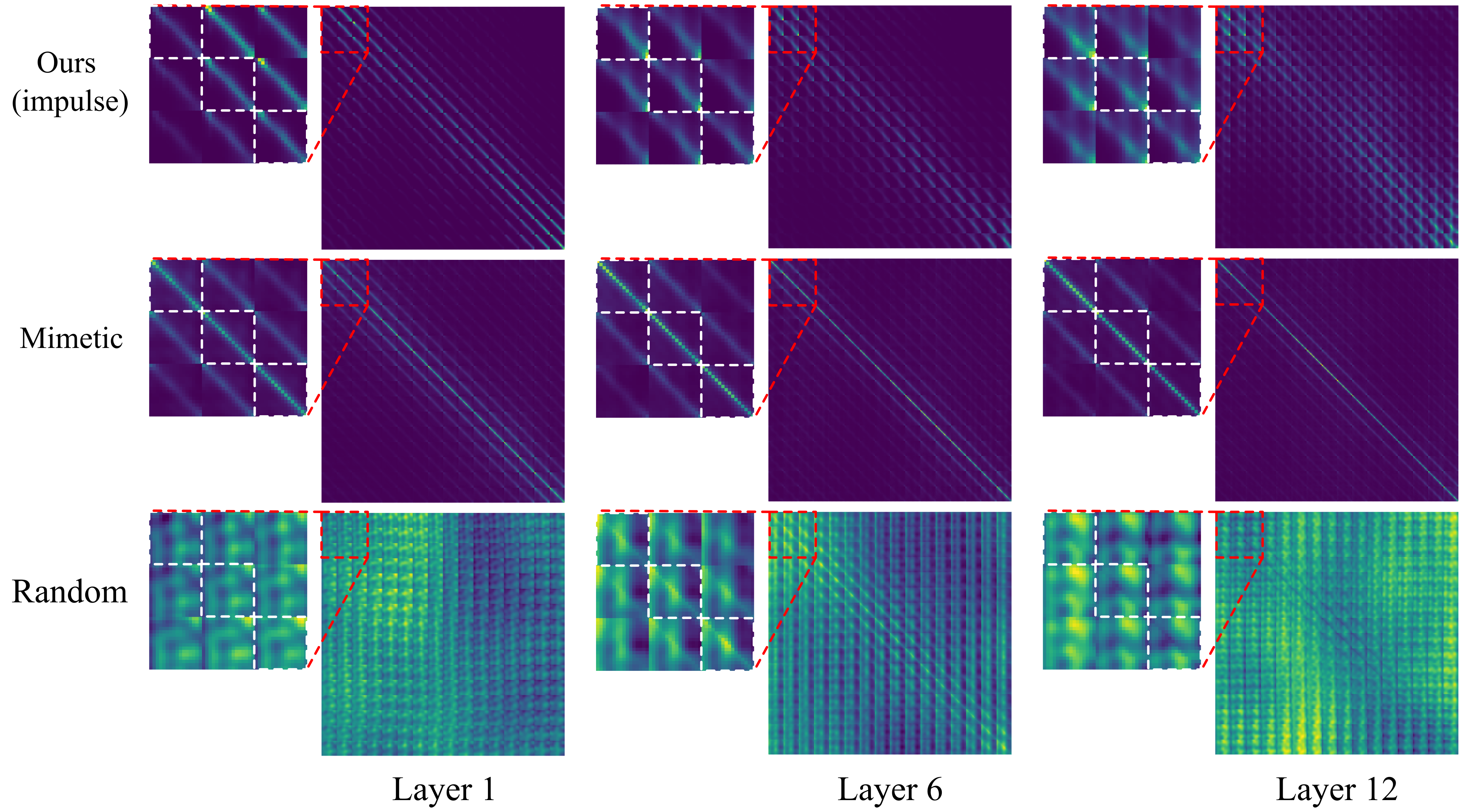}
    \caption{Visualization of attention maps in ViT-T using our impulse initialization method, mimetic~\cite{trockman2023mimetic}, and random~\cite{liu2022convnet} initializations. 
    Red boxes highlight zoomed-in details of the $48{\times}48$ upper left corner in attention maps. 
    White boxes indicate the main diagonal blocks of the zoomed-in attention maps. 
    Our structured initialization method offers off-diagonal attention peaks aligned with the impulse structures, whereas mimetic initialization primarily strengthens the main diagonal of the attention map.
    Random initialization shows little to no patterns.}
    \label{fig:vis_attn}
    \vspace{-0.4cm}
\end{figure}

\vspace{-0.2cm}
\subsection{Attention Maps}
\label{sec:exp_map}
\vspace{-0.2cm}
To show the effectiveness of our initialization on constraining the structure of attention maps, we show the averaged attention maps across CIFAR-10 training data using ViT-T/h3 at 1st, 6th, and 12th layers in~\cref{fig:vis_attn}.
The attention map showed here does not have an exact pattern as seen in impulse filters since we use pseudo inputs to optimize the initial attention parameters.
Nevertheless, there exists a clear pattern of convolutional structures.
As the network layer becomes deeper, this convolutional structure becomes less visible.

\begin{table}[t]
\caption[]{Classification accuracy(\%) of ConvMixer (depth 8) with different filter sizes, embedding dimensions on CIFAR-10.}
\centering
\begin{adjustbox}{width=0.9\linewidth}
\begin{tabular}
{@{}x{0.1\linewidth}cx{0.1\linewidth}x{0.1\linewidth}x{0.1\linewidth}x{0.1\linewidth}cx{0.1\linewidth}x{0.1\linewidth}x{0.1\linewidth}x{0.1\linewidth}@{}}
\toprule 
\multirow{2}{*}{\begin{tabular}[c]{@{}c@{}}\thead{\normalsize Kernel\\ Size}\end{tabular}} 
&& \multicolumn{4}{c}{Embedding Dimension $=$ 256} 
&& \multicolumn{4}{c}{Embedding Dimension $=$ 512} \\
\cmidrule{3-6}\cmidrule{8-11} 
& & Trained & Random & Impulse & Box
& & Trained & Random & Impulse & Box\\
\midrule
3 && 91.76 & 90.72 & 90.68 & 81.70 && 92.82 & 92.15 & 92.20 & 81.90 \\
5 && 92.69 & 90.87 & 90.41 & 80.57 && 93.90 & 92.72 & 91.91 & 81.19 \\
8 && 92.34 & 88.12 & 87.82 & 78.95 && 92.96 & 90.09 & 89.61 & 80.10\\
\bottomrule
\end{tabular}
\label{tab:conv}
\end{adjustbox}
\vspace{-0.4cm}
\end{table}

\vspace{-0.2cm}
\subsection{Comparing ViT with ConvMixer}
\label{sec:exp_convmixer}
\vspace{-0.2cm}
To validate our findings discussed in~\cref{sec:sm_conv} regarding the effectiveness of random filters, we train ConvMixer~\cite{trockman2022patches} models with an embedding dimension of 256, a depth of 8, and a patch size of 2 on the CIFAR-10 dataset, using filter size of 3, 5, and 8.
We follow the same configurations as defined in~\cref{sec:exp_settings}, except for setting the learning rate to 0.01, and the number of epochs to 100. 
Additionally, we tested with a version of ConvMixer with an embedding dimension of 512. 
The results are shown in~\cref{tab:conv}.

We tested on the end-to-end trained ConvMixer along with three different initialization methods: random, impulse, and box.
Please note that the three initialization methods only initialize the spatial convolution filters without training.
Specifically, the box filters use all ones, serving as an average pooling function.

In general, random and impulse initialization achieve comparable accuracy compared to the end-to-end trained model, while box initialization exhibits inferior performance.
This discrepancy can be attributed to the deficient rank of box filters, as they lack $f^2$ linearly independent filters, unlike random and impulse initialization, which can form the basis of the filter space.

For ConvMixers with the same embedding dimension, the performance gap between trained filters and random or impulse widens as the kernel size increases. 
As we discussed in~\cref{sec:exp_larger}, as the kernel size increases, more unique filters are needed to form the basis of filter space.
Consequently, each head (unique filter) has fewer copies, making it difficult to match the rank of inputs, thus failing to meet the condition in~\cref{equ:conv_condition}.
When the embedding dimension doubles, the performance gap between trained filters and random or impulse filters diminishes with the same kernel size. 
However, models with larger kernel sizes still tend to have a bigger gap due to an insufficient number of copies for each unique filter.

Our method is motivated by the similarity between ViT and ConvMixer. 
To show this connection, we train ViTs with similar configurations to ConvMixer as described in~\cref{sec:exp_convmixer}. 
Specifically, we train ViTs of a depth of 8 with embedding dimensions of 256 and 512. 
The number of heads is from 4 to 32. 
Results of different initialization methods are shown in~\cref{tab:vit_conv}.

Our impulse initialization methods demonstrate superior performance across nearly all configurations.
Especially, our method achieves a top accuracy of $90.67\%$ and $91.87\%$ with an embedding dimension of 256 and 512, respectively, significantly outperforming other initialization methods.
Moreover, our method achieves results on par with end-to-end trained ConvMixers ($91.76\%$ and $92.82\%$) of equal depth and embedding dimensions.
In addition, these results with different numbers of heads validate our discussions in~\cref{sec:exp_larger} regarding the impact of the number of heads on the performance.

\begin{table}[t]
\caption[]{Classification accuracy(\%) of ViT (depth 8) with a different number of heads, embedding dimensions on CIFAR-10.}
\centering
\begin{adjustbox}{width=0.9\linewidth}
\begin{tabular}
{@{}llx{0.09\linewidth}x{0.09\linewidth}x{0.09\linewidth}x{0.09\linewidth}cx{0.09\linewidth}x{0.09\linewidth}x{0.09\linewidth}x{0.09\linewidth}@{}}
\toprule 
&\multirow{2}{*}{\thead{\normalsize Method}} 
& \multicolumn{4}{c}{Embedding Dimension $=$ 256} 
&& \multicolumn{4}{c}{Embedding Dimension $=$ 512} \\
\cmidrule{3-6}\cmidrule{8-11} 
& & h4 & h8 & h16 &h32
& & h4 & h8 & h16 &h32\\

\midrule
&Kaiming Uniform~\cite{he2015delving} 
& 85.71 & 85.16 & 84.62 & 84.24 && 87.28 & 86.50 & 85.07 & 84.17 \\
&Trunc Normal 
& 87.27 & 87.10 & 87.30 & 86.71 && 87.49 & 87.03 & 87.74 & 87.39 \\
&Mimetic~\cite{trockman2023mimetic} 
& \textbf{90.52} & 89.45 & 88.97 & 86.83 && 90.94 & 90.75 & 90.35 & 89.27\\
&Ours (Imp.-3)  
& 89.95 & \textbf{90.67} & \underline{90.59} & \underline{88.69} && \textbf{91.55} & \textbf{91.75} & \underline{91.49} & \textbf{91.18}\\
&Ours (Imp.-5)  
& \underline{90.08} & \underline{90.38} & \textbf{90.61} & \textbf{89.14} && \underline{90.92} & \underline{91.67} & \textbf{91.87} & \underline{90.84}\\
\bottomrule
\end{tabular}
\label{tab:vit_conv}
\end{adjustbox}
\vspace{-0.4cm}
\end{table}

\vspace{-0.2cm}
\section{Limitations}
\vspace{-0.2cm}
\textbf{(1) Pseudo input.}\,\,
Although positional encoding is a suitable choice for pseudo input considering its simplicity and its data-independent nature, there may exist even simpler choices for the pseudo input.
Since PE does not account for network depth, the actual attention maps shown in~\cref{fig:vis_attn} appear more blurred as the network depth increases.

\noindent\textbf{(2) Pre-optimization step.}\,\,
The iterative optimization method for solving initial attention parameters is simple and fast, surpassing SVD in mitigating the low-rank approximation errors.
However, leveraging the nonlinearity of the softmax function during initialization may cause network weights to enter a gradient plateau, hindering training in subsequent tasks such as classification.

\noindent\textbf{(3) Hard constraints.}
Our~\cref{prop:convmixer} is based on the presumption that the filters can at least form the basis of filter space, a characteristic inherent in CNNs. 
However, in ViTs, the limited number of heads may be inadequate to span the filter space of a small kernel. 
Finding better adaptions in this scenario remains a challenge.

\noindent\textbf{(4) Value initialization.}
Our method does not consider the initialization for the value weights $\V$ in ViT.

\vspace{-0.2cm}
\section{Conclusion}
\vspace{-0.2cm}
In this paper, we propose a new ViT initialization strategy---convolutional structured impulse initialization---to address the problem that ViTs are difficult to train on small-scale datasets. 
Our initialization requires no off-line knowledge of pre-trained models on large-scale datasets (mimetic or empirical). 
Our strategy is instead inspired by the architectural inductive bias of convolution, without the need for an architectural modification for ViT. 
Unlike traditional generative initialization strategies focusing on the distribution of parameters, our method constrains the structure of attention maps within ViT.
After a careful study of why random spatial convolution filters work, we opt to initialize self-attention maps as random impulse convolution filters, which reinterpret the architectural inductive bias in CNNs as an initialization bias and preserve the architectural flexibility of transformers. 
We validate our methods across various datasets, achieving state-of-the-art performance for ViT trained on small-scale datasets.

\appendix

\section{Convolutional Represetation Matrix}
\label{sec:supp_conv}

In~\cref{sec:sm_conv}, we interchangeably use the terms convolution filter $\mathbf{h}$ and convolution matrix $\mathbf{H}$. 
Additionally, we represent the impulse filter as a convolutional matrix. 
Here, we offer a detailed explanation of the relationship between the convolutional filters and the convolutional matrices.

Let us define a 2D convolution filter as~$\h \,{\in}\, \mathbb{R}^{f {\times} f}$ with elements
\begin{equation}
    \h = 
  \left(\begin{matrix}
  h_{11} & \cdots & h_{1f} \\
  \vdots & \ddots & \vdots \\
  h_{f1} & \cdots & h_{ff}
  \end{matrix}\right).
    \label{equ:supp_h}
\end{equation}
When $\h$ is convolved with an image $\x\,{\in}\,\mathbb{R}^{H {\times} W}$, this convolution operation is equivalent to a matrix multiplication
\begin{equation}
    \mbox{vec}(\h \,{*}\, \x) \,{=}\, \H \, \mbox{vec}(\x),
    \label{equ:supp_conv_01}
\end{equation}
where $\H$ is composed from the elements in $\h$ and zeros in the following format:
\begin{equation}
    \H = 
  \left(\begin{matrix}
  \mathbf{F_1} & \mathbf{F_2} & \cdots & \mathbf{F_f} & \mathbf{0} & \mathbf{0} & \cdots & \mathbf{0} \\
  \mathbf{0} & \mathbf{F_1} & \mathbf{F_2} & \cdots & \mathbf{F_f} & \mathbf{0} & \cdots & \mathbf{0} \\
  \vdots & \ddots & \ddots& \ddots& \ddots& \ddots& \ddots & \vdots \\
  \mathbf{0} & \cdots & \mathbf{0} & \mathbf{F_1} & \mathbf{F_2} & \cdots & \mathbf{F_f} & \mathbf{0} \\
  \mathbf{0} & \cdots & \mathbf{0} & \mathbf{0} & \mathbf{F_1} & \mathbf{F_2} & \cdots & \mathbf{F_f} 
  \end{matrix}\right),
    \label{equ:supp_conv_02}
\end{equation}
where 
\begin{equation}
    \mathbf{F_i} = 
  \left(\begin{matrix}
  h_{i1} & h_{i2} & \cdots & h_{if} & 0 & 0 & \cdots & 0 \\
  0 & h_{i1} & h_{i2} & \cdots & h_{if} & 0 & \cdots & 0 \\
  \vdots & \ddots & \ddots& \ddots& \ddots& \ddots& \ddots & \vdots \\
  0 & \cdots & 0 & h_{i1} & h_{i2} & \cdots & h_{if} & 0 \\
  0 & \cdots & 0 & 0 & h_{i1} & h_{i2} & \cdots & h_{if}
  \end{matrix}\right),
    \label{equ:supp_conv_03}
\end{equation}
for $i\,{=}\,1,2,\,{\dots}\,,f$. 
$\mathbf{F_i}$s are circulant matrices and $\H$ is a block circulant matrix with circulant block (BCCB). 
Note that convolutions may employ various padding strategies, but the circulant structure remains consistent. 
Here, we show the convolution matrix without any padding as an example.

\section{Additional Results}
\label{sec:supp_exp}

\begin{table}[t]
\caption[]{Classification accuracy(\%) of different pseudo input on CIFAR-10.
{\color{tabgreen}Green} shaded row indicates the best choice of pseudo inputs when achieving the best average accuracy.
``PE'' denotes positional encoding, and ``G'' represents random sampling from the Gaussian distribution.
``Imp.-3'' represents a $3{times}3$ impulse filter, and ``Imp.-5'' denotes a $5{\times}5$ impulse filter.
}
\centering
\begin{adjustbox}{width=\linewidth}
\begin{tabular}
{@{}x{0.08\linewidth}x{0.12\linewidth}cx{0.09\linewidth}x{0.09\linewidth}cx{0.09\linewidth}x{0.09\linewidth}cx{0.09\linewidth}x{0.09\linewidth}cx{0.09\linewidth}x{0.09\linewidth}cx{0.09\linewidth}@{}}
\toprule
\multicolumn{2}{c}{\thead{\normalsize Pseudo Input}} 
& \multicolumn{6}{c}{\thead{\normalsize Same $\Q_{\text{init}}^{\hphantom{T}}$ and $\K_{\text{init}}^{\hphantom{T}}$}} 
& \multicolumn{6}{c}{\thead{\normalsize Different $\Q_{\text{init}}^{\hphantom{T}}$ and $\K_{\text{init}}^{\hphantom{T}}$}} 
&& \multirow{3}{*}{\thead{\normalsize Avg.}} \\ 
\cmidrule{1-2}\cmidrule{4-8}\cmidrule{10-14}
\multirow{2}{*}{\begin{tabular}[c]{@{}c@{}}First\\ Layer\end{tabular}} 
& \multirow{2}{*}{\begin{tabular}[c]{@{}c@{}}Following\\ Layers\end{tabular}} 
&& \multicolumn{2}{c}{ViT-T/h3} 
&& \multicolumn{2}{c}{ViT-T/h8} 
&& \multicolumn{2}{c}{ViT-T/h3} 
&& \multicolumn{2}{c}{ViT-T/h8} 
& \\
\cmidrule{4-5}\cmidrule{7-8}\cmidrule{10-11}\cmidrule{13-14}
& & &Imp.-3 & Imp.-5 && Imp.-3 & Imp.-5 && Imp.-3 & Imp.-5 && Imp.-3 & Imp.-5 & \\ 
\hline
\rowcolor{tabgreen!30!}
PE   & PE   && 90.75 & 90.22 && 90.39 & 90.24 && 89.90 & 90.18 && 90.19 & 91.24 && \textbf{90.39} \\
\hline
PE   & U   && 86.90 & 86.10 && 87.99 & 87.86 && 87.35 & 85.56 && 86.40 & 86.61 && 86.85 \\
PE   & PE+U && 89.62 & 88.40 && 89.61 & 89.50 && 88.99 & 89.29 && 89.05 & 89.47 && 89.24 \\
U   & PE   && 90.34 & 89.21 && 90.96 & 89.83 && 90.76 & 90.00 && 91.20 & 90.34 && 90.33 \\
U   & U   && 86.05 & 86.50 && 86.13 & 86.44 && 87.09 & 87.22 && 86.03 & 86.13 && 86.45 \\
U   & PE+U && 89.91 & 89.99 && 89.96 & 89.93 && 89.94 & 89.89 && 89.60 & 89.07 && 89.79 \\
PE+U & PE   && 90.07 & 89.19 && 90.31 & 89.56 && 90.03 & 90.72 && 90.40 & 90.76 && 90.13 \\
PE+U & U   && 86.07 & 86.52 && 85.55 & 85.94 && 86.02 & 86.07 && 86.52 & 85.55 && 86.02 \\
PE+U & PE+U && 89.63 & 89.33 && 90.02 & 88.92 && 89.52 & 89.28 && 89.29 & 89.33 && 89.41 \\
\bottomrule
\end{tabular}
\label{tab:supp:pseudo_input}
\end{adjustbox}
\end{table}

\begin{figure}[t!]
    \centering
    \includegraphics[width=\linewidth]{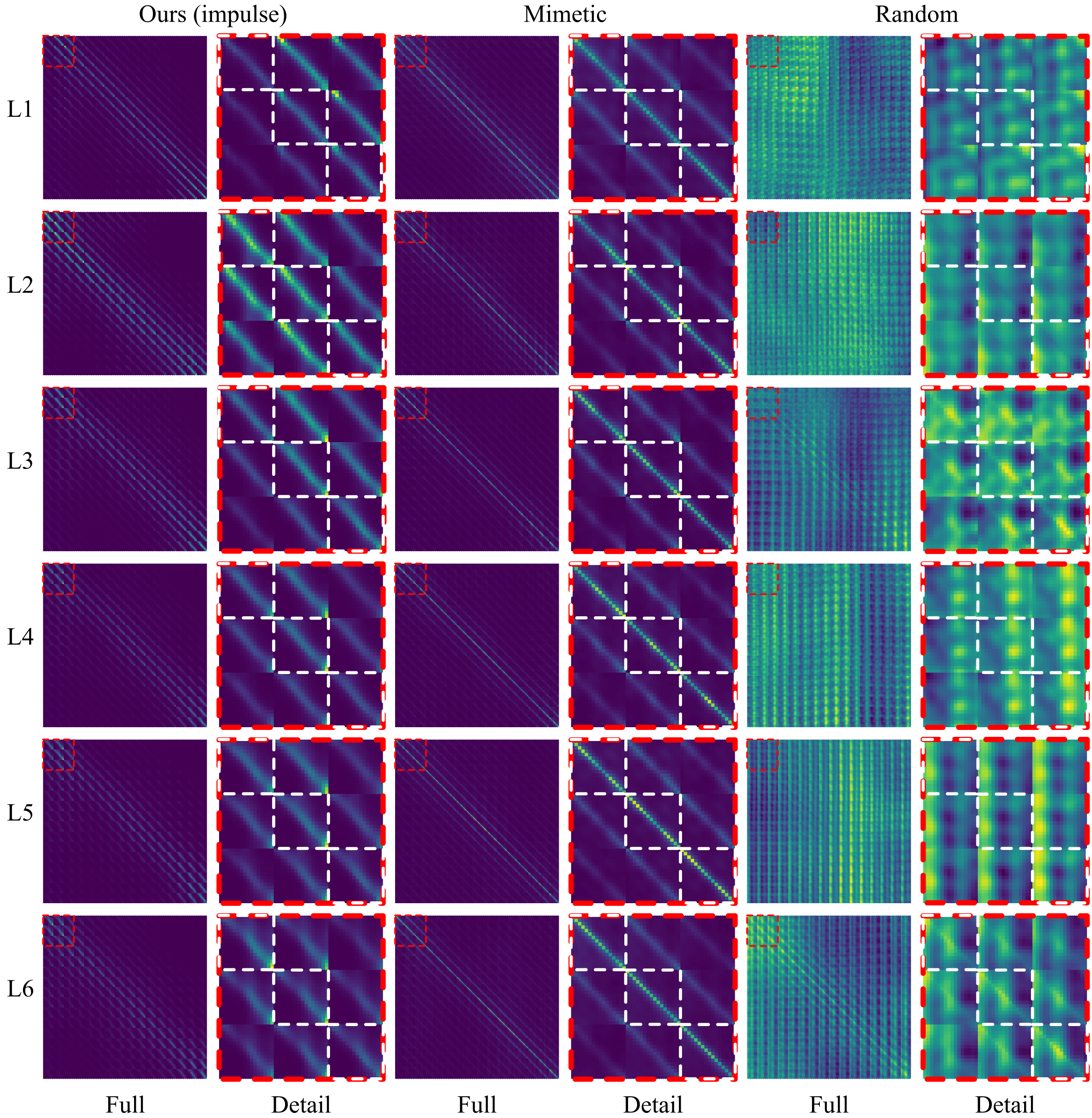}
    \caption{Visualization of attention maps in ViT-T using our impulse initialization method, mimetic~\cite{trockman2023mimetic}, and random~\cite{liu2022convnet} initializations. 
    Red boxes highlight zoomed-in details of the $48{\times}48$ upper left corner in attention maps. 
    White boxes indicate the main diagonal blocks of the zoomed-in attention maps. 
    Our structured initialization method offers off-diagonal attention peaks aligned with the impulse structures, whereas mimetic initialization primarily strengthens the main diagonal of the attention map.
    Random initialization shows little to no patterns.}
    \label{supp:fig:vis_attn_01}
\end{figure}

\begin{figure}[t!]
    \centering
    \includegraphics[width=\linewidth]{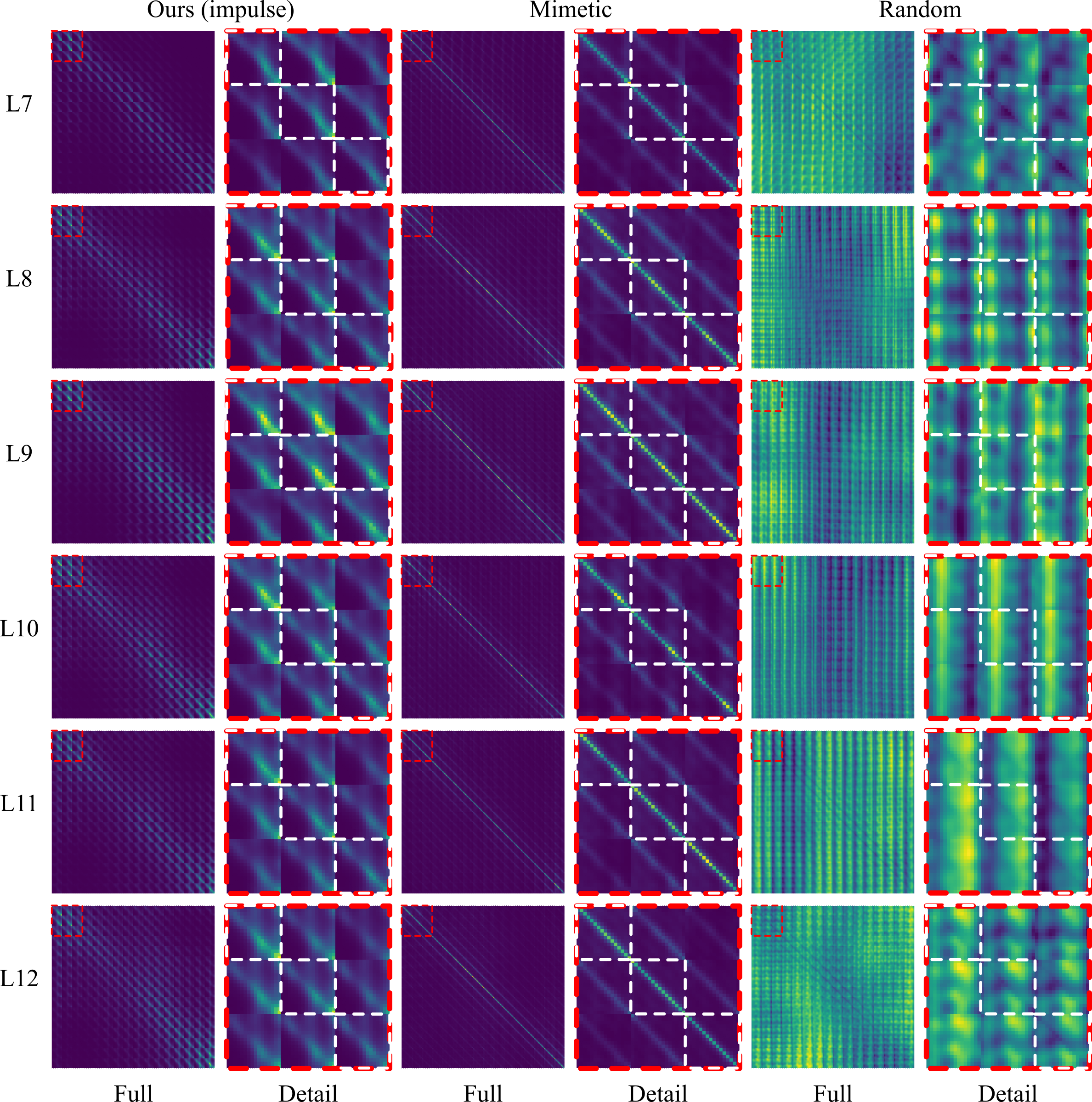}
    \caption{Visualization of attention maps in ViT-T using our impulse initialization method, mimetic~\cite{trockman2023mimetic}, and random~\cite{liu2022convnet} initializations. 
    Red boxes highlight zoomed-in details of the $48{\times}48$ upper left corner in attention maps. 
    White boxes indicate the main diagonal blocks of the zoomed-in attention maps. 
    Our structured initialization method offers off-diagonal attention peaks aligned with the impulse structures, whereas mimetic initialization primarily strengthens the main diagonal of the attention map.
    Random initialization shows little to no patterns.}
    \label{supp:fig:vis_attn_02}
\end{figure}

\begin{table}[t!]
\caption[]{Classification accuracy(\%) of ViT (depth 8) with a different number of heads, embedding dimensions on CIFAR-10.}
\centering
\begin{adjustbox}{width=0.9\linewidth}
\begin{tabular}
{@{}llx{0.09\linewidth}x{0.09\linewidth}x{0.09\linewidth}x{0.09\linewidth}cx{0.09\linewidth}x{0.09\linewidth}x{0.09\linewidth}x{0.09\linewidth}@{}}
\toprule 
&\multirow{2}{*}{\thead{\normalsize Method}} 
& \multicolumn{4}{c}{Embedding Dimension $=$ 64} 
&& \multicolumn{4}{c}{Embedding Dimension $=$ 128} \\
\cmidrule{3-6}\cmidrule{8-11} 
& & h4 & h8 & h16 &h32
& & h4 & h8 & h16 &h32\\
\midrule
&Kaiming Uniform~\cite{he2015delving} 
& 81.26 & 80.76 & 80.71 & \underline{79.89} && 84.46 & 83.83 & 83.05 & 82.70 \\
&Trunc Normal 
& 81.57 & 82.12 & 81.57 & \textbf{80.40} && 86.19 & 86.00 & 85.39 & 84.37 \\
&Mimetic~\cite{trockman2023mimetic} 
& \textbf{84.59} & \underline{82.78} & \textbf{81.78} & 79.71 && \underline{87.90} & 87.49 & 85.51 & 83.68\\
&Ours (Imp.-3)  
& 82.73 & \textbf{84.12} & \underline{81.59} & 79.39 && \textbf{88.49} & \underline{87.75} & \textbf{87.63} & \underline{84.42}\\
&Ours (Imp.-5)  
& \underline{83.43} & 82.66 & 80.80 & 79.11 && 87.82 & \textbf{88.02} & \underline{87.57} & \textbf{84.79}\\
\bottomrule
\end{tabular}
\label{tab:supp_vit_conv}
\end{adjustbox}
\end{table}

\subsection{Pseudo Input}
\label{sec:supp_exp_input}
We offer additional results for the ablation study on pseudo input with random inputs sampled from a Uniform distribution in~\cref{tab:supp:pseudo_input}. 
The trend is aligned with the findings in the main paper for random inputs sampled from a Gaussian distribution.
However, the performance of using random inputs sampled from a Uniform distribution is worse than those using a Gaussian distribution, since the real input data is more likely to follow a Gaussian distribution.

\subsection{Relationship between Head Numbers and Embedding Dimension}
\label{sec:supp_exp_convmixer}

In the main paper, we have discussed the relationship between the number of heads and the embedding dimensions. 
We have provided the results with embedding dimensions of 256 and 512 for a ViT with depth 8. 
Here we show additional results with embedding dimensions of 64 and 128 in~\cref{tab:supp_vit_conv}. 
The trend is consistent with the discussions in the main paper.

\subsection{Attention Maps}
\label{sec:supp_exp_map}

Here we provide additional visualization of the attention maps for all 12 layers in~\cref{supp:fig:vis_attn_01} and~\cref{supp:fig:vis_attn_02}.

%
%
\bibliographystyle{splncs04}
\bibliography{main}
\end{document}